\documentclass[
]{ceurart}

\begin{document}

\copyrightyear{2021}
\copyrightclause{Copyright for this paper by its authors.
  Use permitted under Creative Commons License Attribution 4.0
  International (CC BY 4.0).}

\conference{FIRE 2021: Forum for Information Retrieval Evaluation, India}

\title{Findings of the Sentiment Analysis of Dravidian Languages in Code-Mixed Text}

\author[1]{Bharathi Raja Chakravarthi}[%
orcid=0000-0002-4575-7934,]
\ead{bharathi.raja@insight-centre.org}
\address[1]{Insight SFI Research Centre for Data Analytics, Data Science Institute, National University of Ireland Galway}

\author[2]{Ruba Priyadharshini}[orcid=0000-0003-2323-1701]
\address[2]{ULTRA Arts and Science College, Madurai, Tamil Nadu, India}
\ead{rubapriyadharshini.a@gmail.com}

\author[3]{Sajeetha Thavareesan}[orcid=0000-0002-6252-5393]
\address[3]{Eastern University, Sri Lanka}
\ead{sajeethas@esn.ac.lk}

\author[4]{Dhivya Chinnappa}[orcid=0000-0002-8318-3337]
\ead{dhivya.infant@gmail.com}
\address[4]{Thomson Reuters, USA}

\author[5]{Durairaj Thenmozhi}[orcid=]
\ead{theni\_d@ssn.edu.in}
\address[5] {Sri Sivasubramaniya Nadar College of Engineering, Tamil Nadu, India}

\author[6]{Elizabeth Sherly}[orcid=0000-0001-6508-950X]
\ead{sherly@iiitmk.ac.in}
\address[6]{Indian Institute of Information Technology and Management-Kerala, India}

\author[1]{John P. McCrae}[orcid=0000-0002-7227-1331]
\ead{john.mccrae@insight-centre.org}

\author[8]{Adeep Hande}[orcid=0000-0002-2003-4836]
\ead{adeeph18c@iiitt.ac.in}
\address[8]{Indian Institute of Information Technology Tiruchirappalli}

\author[6]{Rahul Ponnusamy}[orcid=0000-0001-8023-7742]
\ead{rahul.mi20@iiitmk.ac.in}

\author[10]{Shubhanker Banerjee}[orcid=0000-0002-3969-5183]
\ead{S.Banerjee3@nuigalway.ie}
\address[10]{ADAPT Centre, Data Science Institute, National University Of Ireland Galway}
\author[11]{Charangan Vasantharajan}[orcid=0000-0001-7874-3881]
\ead{charangan.18@cse.mrt.ac.lk}
\address[11]{Department of Computer Science and Engineering, University of Moratuwa, Sri Lanka}
\begin{abstract}
We present the results of the Dravidian-CodeMix shared task\footnote{\url{https://dravidian-codemix.github.io/2021/index.html/}} held at FIRE 2021, a track on sentiment analysis for Dravidian Languages in Code-Mixed Text. We describe the task, its organization, and the submitted systems. This shared task is the continuation of last year's Dravidian-CodeMix shared task\footnote{\url{https://dravidian-codemix.github.io/2020/index.html/}} held at FIRE 2020. 
This year's tasks included code-mixing at the intra-token and inter-token levels. 
Additionally, apart from Tamil and Malayalam, Kannada was also introduced.
We received 22 systems for Tamil-English, 15 systems for Malayalam-English, and 15 for Kannada-English.  The top system for Tamil-English, Malayalam-English and Kannada-English scored weighted average F1-score of 0.711, 0.804, and 0.630, respectively. In summary, the quality and quantity of the submission show that there is great interest in Dravidian languages in code-mixed setting and state of the art in this domain still needs more improvement. 
\end{abstract}


\maketitle

\section{Introduction}
Sentiment analysis is a text mining task that finds and extracts personal information from the source material, allowing a company/researcher to understand better the social sentiment of its brand, product, or service while monitoring online conversations \cite{Hande2021BenchmarkingML}. 
In our case, we used the comments from the movie trailer, so it is about finding the viewers sentiment of the movie. The constantly increasing number of social media and user-generated comments raises the importance of finding sentiments in local languages as making these predictions is essential for local businesses. For this study, we created data in Dravidian languages, namely Tamil (ISO 639-3:tam), Malayalam (ISO 639-3:mal), and Kannada (ISO 639-3:kan). Tamil is the official language of Tamil Nadu, the Indian Union, Sri Lanka, Malaysia and is spoken in many places in South Asian countries. Malayalam and Kannada have official status in the Indian Union government and the state of Kerala and Karnataka, respectively \cite{Chakravarthi2021DatasetFI,hande-etal-2020-kancmd}. 

The Tamil script, the Vatteluttu alphabet, and the Chola-Pallava script all came together to form the Tamil script. The Tamil script dates back to 600 BCE, found at various archaeological sites in Tamil Nadu, Sri Lanka, Egypt, Thailand, Vietnam, Cambodia and Indonesia.  The Chola-Pallava script is the ancestor of the present Tamil script. Thani Tamil Iyakkam (Pure or Independent Tamil Movement) is a Tamil linguistic purity movement that tried to avoid borrowing terms from Sanskrit, English, and other languages in 1916. Maraimalai Adigal\footnote{\url{https://en.wikipedia.org/wiki/Maraimalai_Adigal}}, Paventhar Bharathidasan\footnote{\url{https://en.wikipedia.org/wiki/Bharathidasan}}, Devaneya Pavanar\footnote{\url{https://en.wikipedia.org/wiki/Devaneya_Pavanar}}, and Pavalareru Perunchitthiranaar\footnote{\url{https://en.wikipedia.org/wiki/Perunchithiranar_(Tamil_nationalist)}} started the movement, which was spread through the Thenmozhi literary journal created by Pavalareru P. The natural continuation of this endeavour was to purge Tamil of Sanskrit influence including negative societal attitudes such  discrimination based on colour and birth, central discrimination being education only for particular people which denies education for the main population that they felt kept Tamils in a condition of economic, cultural, and political slavery, which they believed made Tamil and other Dravidian states vulnerable to external political dominance.

Despite the vast amounts of primary and secondary speakers, Kannada is a low resource language concerning language technology. It primarily speaks by people in Karnataka, India, and is also the state's official language. Catanese, the Kannada script, is an alpha-syllabary of the scripts of the Brahmic family evolving
into the Kadamba script and used to write
other under-resourced languages like Tulu, Konkani, and Sankethi. The Kannada script has 13 vowels (14 if the obsolete vowel includes), 34 consonants, and 2 yogavahakas (semiconsonants: part-vowel, part-consonant). Malayalam used Vatteluttu script and Pallava-Grantha script.   However, by 2020 language mixing of foreign languages in the Dravidian language has become very frequent. English is seen as a predominant language economically and culturally by Dravidian languages speakers, so social media users often adopted Roman script and mixed native script. 

The Dravidian-CodeMix task was introduced in 2020 and aimed to explore the sentiment analysis of code mixed comments in Dravidian languages. In 2020, we released the data for Tamil and Malayalam in Roman script.  The dataset included 15,000 instances for Tamil and 6,000 instances for Malayalam. In 2021, apart from Tamil and Malayalam, we introduce a Sentiment Analysis dataset for Kannada 
Thus, in 2021 we will include three languages Tamil, Malayalam, and Kannada.
Our dataset contains all kinds of code-mixing, ranging from simple script mixing to mixing at the morphological level. 
The challenge is to determine the polarity of sentiment in a code-mixed dataset of comments or posts in Tamil-English, Malayalam-English, and Kannada-English \cite{Hande2021OffensiveLI}. More details about the annotation of the dataset can be found in \cite{hande-etal-2020-kancmd,chakravarthi-etal-2020-corpus, chakravarthi-etal-2020-sentiment}  

Our shared task seeks to promote a study on how sentiment communicates on Dravidian social media language in a code-mixed setting and aim for better social media analysis \cite{chakravarthi-2020-hopeedi,Hande2021HopeSD}. 
We presented the training, development and test set to the participants.
This paper presents an overview of the task description, dataset, description of the participating systems, analysis, and provide insights from the shared task.
 
\section{Task Description}

This task aims at the classification of sentence-level polarities. The main objective of the proposed systems is to classify the polarity of a given YouTube comment into mixed feelings, negative and positive or identify if the given comment does not belong to one of the following languages of this shared task: Tamil-English, Malayalam-English, and Kannada-English.
The comments provided to the participants were written in a mixture of Latin script, native script, and both Latin script and native script. Some of the comments followed the grammar of one of the Dravidian languages: Tamil or Malayalam, or Kannada, but are written using the English lexicon. Other comments followed the lexicon of the Dravidian languages and were written using English grammar.
The participants were provided with the development, training and test dataset. This is a message-level polarity classification task. Participants' systems have to classify a Youtube comment into positive, negative, neutral, mixed emotions, or not in the intended languages. 

The following examples are from the Tamil dataset illustrate dataset code-mixing.

\begin{itemize}
    \item {\color{blue}\textbf{Epo pa varudhu indhe padam} }\-- \textit{When will this movie come out?} Tamil words written in Roman script with no English switch.
    \item \textbf{{\color{blue}{Yaru viswasam \color{red}{teaser} \color{blue}{ku marana} \color{red}{waiting like}  \color{blue}{pannunga}}}} \-- \textit{Who is waiting for Viswarm teaser, please like} Tag switching with English words.
    \item \textbf{{\color{red}Omg .. use head phones. }{\color{blue}Enna {\color{red}bgm} da saami ..}} \-- \textit{OMG! Use your headphones. Good Lord, What a background score!} Inter-sentential switch
    \item \textbf{{\color{red}I think {\color{blue}sivakarthickku} hero getup set {\color{blue}aagala.}}} \-- \textit{I think the hero role does not suit Sivakarthick.} Intra-sentential switch between clauses.
\end{itemize}

The following examples are from the Malayalam dataset.

\begin{itemize}
    \item {\color{blue}\textbf{Orupaadu nalukalku shesham aanu ithupoloru padam eranghunnathu.} }\-- \textit{A movie like this is coming out after a long time.} Malayalam words written in Roman script with no English switch.
    \item {\color{blue}\textbf{{\color{red}Malayalam industry} ku {\color{red}thriller} kshamam illannu kaanichu kodukku anghotu.}} \-- \textit{Show that there is no shortage for thriller movies in Malayalam film industry.} Tag switching with English words.
    \item \textbf{{\color{blue}Manju chechiyude athyugran {\color{red}performance}nayi kaathirikunnu.} {\color{red}  The Lady superstar of Malayalam industry.}}  \-- \textit{Waiting for the awesome performance of Manju sister. The Lady superstar of Malayalam film industry.} Inter-sentential switch
    \item \textbf{{\color{red}Next movie ready for {\color{blue}nammude swantham dhanush.} }} 
    \-- \textit{Next movie ready for our dear Dhanush.}  
 
    \item {\color{blue}\textbf{Orupaadu nalukalku shesham aanu ithupoloru padam eranghunnathu.} }\-- \textit{A movie like this is coming out after a long time.} Malayalam words are written in Roman script with no English switch.
    
\end{itemize}
  
The following examples are from the Kannada dataset.
\begin{itemize}
    \item  \textbf{\textcolor{blue}{Yaru} \textcolor{red}{tension} \textcolor{blue}{agbede yakandre} \textcolor{red}{dislike} \textcolor{blue}{madiravru mindrika kadeyavru}} --\textit{ No one needs to worry as the people who disliked this are fans of Mandrika}. Intra-sentential switch between clauses
    
    \item \textbf{\textcolor{blue}{Gottilla Rakshit Shettru natana nanu fida.} \textcolor{red}{Boss waiting for} \textcolor{red}{movie} \textcolor{blue}{Charitre bareyo} \textcolor{blue}{ella lakshana ide.} \textcolor{red}{All the best for you bright future}} --\textit{Don’t know why, I am obsessed with Rakshit Shetty’s acting. waiting for your movie, expecting it to be a blockbuster. All the best for your bright future.} Inter-sentential and intra-sentential mix. (Kannada written in both Latin and Kannada script)
    
    \item \textbf{\textcolor{blue}{Nanage ansutte ee \textcolor{red}{video} vanna rashmika mandanna} \textcolor{red}{fans dislike} \textcolor{blue}{madirbahudu}} --\textit{I feel that this video has been disliked by the fans of Rashmika Mandana.}Intra-sentential switch between clauses.Code-switching at morphological level: (written in both Kannada and Latin script)
    
    \item \textbf{\textcolor{red}{My favorite song in 2019 is Taaja Samachara.} \textcolor{blue}{Sahitya priyare omme ee haadu kelidre kelthane irabeku ansutte.} \textcolor{red}{Everybody watch this.}} --\textit{My favourite song in 2019 is Taaja Sanachara. Literature admirers, please listen to the song once; you would want to listen to it over and over again. Everybody watch this.} Inter-sentential code-mixing: Mix of English and Kannada (Kannada written in Kannada script itself)
\end{itemize}
 
The data was annotated for sentiments according to the following schema. 
\begin{itemize}
    \item \textbf{Positive state:} The text contains an explicit or implicit indication that the speaker is in an optimistic mood, i.e., joyful, admiring, relaxed, and forgiving. 
    \item \textbf{Negative state:} The text contains an explicit or implicit indication that the speaker is in an unfavourable condition, i.e., depressed, angry, nervous, or aggressive. 
    \item \textbf{Mixed feelings:} The text contains an explicit or implicit indication indicating that the speaker is experiencing both good and negative emotions. Comparing two films  
    \item \textbf{Neutral state:} There is no explicit or implicit indication of the speaker's emotional state: examples include requests for likes or subscriptions, as well as inquiries about the release date or movie dialogue. This is a state that can be termed neutral.
    \item \textbf{Not in intended language:} For Kannada, if the sentence does not contain Kannada, then it is not Kannada.
\end{itemize}
The annotators were provided with Tamil, Kannada, and Malayalam translations of the above to facilitate better understanding. A minimum of three annotators annotated each sentence. Dataset corpus statistics are given in Table \ref{tab:corp_stat1}, Table \ref{tab:senti-data_distribution}, and Table \ref{tab:train_data_distribution-senti}.

\begin{table*}
\begin{center} 
\renewcommand{\tabcolsep}{1.5mm}

\normalsize
\begin{tabular}{|l|r|r|r|}
\hline
Language & Tamil  &Malayalam &Kannada\\
\hline
Number of words & 513,311 & 224,207 & 65,002\\
Vocabulary size & 94,928 & 57,566 & 20,665\\
Number of comments & 44,020 & 19,616 & 7,671\\
Number of sentences & 52,750 & 24,014 & 8,472\\
Average number of words per sentence & 11 & 11 & 8\\
Average number of sentences per comment & 1 & 1 & 1\\
\hline
\end{tabular} 
\caption{Corpus statistics of the dataset} 
\label{tab:corp_stat1} 
\end{center} 
\end{table*}

\begin{table}[]  
\begin{center} 
\renewcommand{\tabcolsep}{1.5mm}
\normalsize
\begin{tabular}{|l|r|r|r|}
\hline
Class & Tamil & Malayalam & Kannada \\
\hline
Negative & 5,228 (11.87 \%) & 2,600 (13.25 \%) & 1,484 (19.34 \%)\\
Not in intended language & 2,087 (4.74 \%) & 1,445 (736 \%) & 1,136 (14.80 \%)\\
Neutral state & 6,904 (15.68 \%) & 6,502 (33.14 \%) & 842 (10.97 \%)\\
Mixed feelings & 4,928 (1119 \%) & 1,162 (5.92 \%) & 691 (9.00 \%)\\
Positive & 24,873 (56.50 \%) & 7,907 (40.30 \%) & 3,518 (45.86 \%)\\
\hline
Total & 44,020 & 19,616 & 7,671\\
\hline
\end{tabular} 
\caption{Class-wise Dataset Distribution} 
\label{tab:senti-data_distribution} 
\end{center} 
\end{table}

\begin{table}  
\begin{center} 
\renewcommand{\tabcolsep}{1.5mm}
\normalsize
\begin{tabular}{|l|r|r|r|}
\hline
 &   Tamil & Malayalam & Kannada\\
\hline
Training & 35,220 & 15,694 & 6,136\\
Development & 4,398 & 1,960 & 767\\
Test & 4,402 & 1,962 & 768\\
\hline
Total & 44,020 & 19,616 & 7,671\\
\hline
\end{tabular} 
\caption{Train-Development-Test Data Distribution with 90\%-5\%-5\% train-dev-test split} 
\label{tab:train_data_distribution-senti} 
\end{center} 
\end{table}

\begin{table}
\begin{center} 
 \small
\begin{tabular}{|l|l|r|r|r|r|}
\hline
No.  &  TeamName  &  Precision  &  Recall  &  F1-Score  &  Rank\\
\hline

1 & SSNCSE\_NLP \cite{SSNCSE_NLP_aDravidianCodeMix21} & 0.64 & 0.66 & 0.63 & 1 \\
2 & MUCIC \cite{MUCIC} & 0.62 & 0.66 & 0.63 & 2 \\
3 & CIA\_NITT \cite{CIA_NITT} & 0.63 & 0.64 & 0.63 & 3 \\
4 & SOA-NLP \cite{SOA_NLP} & 0.64 & 0.65 & 0.62 & 4 \\
5 & IIITT-Karthik Puranik \cite{IIITT} & 0.62 & 0.63 & 0.62 & 5 \\
6 & Dynamic Duo \cite{DynamicDuo} & 0.67 & 0.65 & 0.62 & 6 \\
7 & KBCNMUJAL \cite{KBCNMUJAL} & 0.62 & 0.64 & 0.62 & 7 \\
8 & IIITT-Pawan \cite{IIIT} & 0.61 & 0.61 & 0.61 & 8 \\
9 & AI ML & 0.62 & 0.60 & 0.61 & 9 \\
10 & SSN\_NLP\_MLRG \cite{SSN_NLP_MLRG} & 0.60 & 0.59 & 0.60 & 10 \\
11 & Amrita\_CEN \cite{Amrita_CEN_NLP} & 0.60 & 0.58 & 0.57 & 11 \\
12 & IIIT\_DWD  & 0.57 & 0.54 & 0.55 & 12 \\
13 & LogicDojo & 0.43 & 0.56 & 0.48 & 13 \\
14 & MUM \cite{MUM} & 0.41 & 0.49 & 0.37 & 14 \\
15 & IRLab@IITBHU \cite{IRLab} & 0.29 & 0.35 & 0.32 & 15 \\

\hline
\end{tabular} 
\caption{Rank list based on weighted average F1-score along with other evaluation metrics (Precision and Recall) for the Kannada track} 
\label{tab:res_kan} 
\end{center} 
\end{table}

\section{Methodology}
We received 54 submissions for the task, out of which 17 were for the Malayalam track, 22 were for the Tamil track, and 15 were for the Kannada track. The rank lists for the Kannada track, Tamil track, and the Malayalam track are shown in Tables \ref{tab:res_kan}, \ref{tab:res_tam} and \ref{tab:res_mal} respectively. Below we briefly describe the systems proposed by the top 3 teams in both tracks.

\begin{table}
\begin{center} 
 \small
\begin{tabular}{|l|l|r|r|r|r|}
\hline
No. &  TeamName & Precision & Recall & F1-Score & Rank\\
\hline
1 & CIA\_NITT \cite{CIA_NITT} & 0.71 & 0.71 & 0.71 & 1 \\
2 & ZYBank-AI \cite{ZYBank-AI} & 0.68 & 0.68 & 0.68 & 2 \\
3 & IIITT-Pawan \cite{IIIT} & 0.62 & 0.65 & 0.63 & 3 \\
4 & IIITT-Karthik Puranik \cite{IIITT} & 0.62 & 0.64 & 0.62 & 4 \\
5 & MUCIC \cite{MUCIC} & 0.61 & 0.64 & 0.62 & 5 \\
6 & SOA\_NLP \cite{SOA_NLP} & 0.61 & 0.65 & 0.62 & 6 \\
7 & Ryzer \cite{RYZER} & 0.60 & 0.61 & 0.60 & 7 \\
8 & SSN\_NLP\_MLRG \cite{SSN_NLP_MLRG} & 0.60 & 0.61 & 0.60 & 8 \\
9 & AIML \cite{AIML} & 0.60 & 0.60 & 0.60 & 9 \\
10 & KBCNMUJAL \cite{KBCNMUJAL} & 0.58 & 0.60 & 0.59 & 10 \\
11 & SSNCSE\_NLP \cite{SSNCSE_NLP_aDravidianCodeMix21} & 0.60 & 0.64 & 0.59 & 11 \\
12 & KonguCSE & 0.57 & 0.62 & 0.57 & 12 \\
13 & MUM \cite{MUM} & 0.58 & 0.62 & 0.56 & 13 \\
14 & LogicDojo  & 0.54 & 0.59 & 0.56 & 14 \\
15 & IIIT DWD \cite{IIIT_DWD} & 0.55 & 0.56 & 0.56 & 15 \\
16 & IIIT Surat \cite{IIITSurat} & 0.54 & 0.57 & 0.55 & 16 \\
17 & Amrita\_CEN \cite{Amrita_CEN_NLP} & 0.64 & 0.50 & 0.53 & 17 \\
18 & SSN-NLP  & 0.62 & 0.49 & 0.51 & 18 \\
19 & DLRF  & 0.34 & 0.58 & 0.42 & 19 \\
20 & IRLab@IITBHU \cite{IRLab} & 0.38 & 0.46 & 0.41 & 20 \\
21 & SSNHacML \cite{SSNHacML} & 0.38 & 0.45 & 0.41 & 21 \\
22 & SSN\_IT\_NLP \cite{SSN_IT_NLP} & 0.38 & 0.39 & 0.38 & 22 \\


\hline
\end{tabular} 
\caption{Rank list based on weighted average F1-score along with other evaluation metrics (Precision and Recall) for the Tamil track} 
\label{tab:res_tam} 
\end{center} 
\end{table}

\begin{table}
\begin{center} 
 \small
\begin{tabular}{|l|l|r|r|r|r|}
\hline
No.  &  TeamName  &  Precision  &  Recall  &  F1-Score  &  Rank\\
\hline

1 & ZYBank-AI Team \cite{ZYBank-AI} & 0.80 & 0.81 & 0.80 & 1 \\
2 & CIA\_NITT \cite{CIA_NITT} & 0.75 & 0.76 & 0.75 & 2 \\
3 & SOA\_NLP \cite{SOA_NLP} & 0.73 & 0.73 & 0.73 & 3 \\
4 & MUCIC \cite{MUCIC} & 0.73 & 0.73 & 0.73 & 4 \\
5 & AIML \cite{AIML} & 0.72 & 0.72 & 0.72 & 5 \\
6 & IIITT-Pawan\cite{IIIT}  & 0.71 & 0.71 & 0.71 & 6 \\
7 & KBCNMUJAL \cite{KBCNMUJAL} & 0.73 & 0.71 & 0.71 & 6 \\
8 & SSN\_NLP\_MLRG \cite{SSN_NLP_MLRG} & 0.70 & 0.71 & 0.70 & 7 \\
9 & SSNCSE\_NLP \cite{SSNCSE_NLP_aDravidianCodeMix21} & 0.69 & 0.69 & 0.69 & 8 \\
10 & Dynamic Duo \cite{DynamicDuo} & 0.69 & 0.70 & 0.69 & 9 \\
11 & IIITT-Karthik Puranik \cite{IIITT} & 0.65 & 0.67 & 0.65 & 10 \\
12 & IRLab@IITBHU \cite{IRLab} & 0.65 & 0.67 & 0.65 & 10 \\
13 & Amrita\_CEN \cite{Amrita_CEN_NLP} & 0.64 & 0.61 & 0.62 & 11 \\
14 & IIIT DWD \cite{IIIT_DWD} & 0.62 & 0.62 & 0.62 & 12 \\
15 & IIIT Surat \cite{IIITSurat} & 0.63 & 0.63 & 0.61 & 13 \\
16 & MUM \cite{MUM} & 0.62 & 0.63 & 0.60 & 14 \\
17 & LogicDojo & 0.52 & 0.58 & 0.55 & 15 \\

\hline
\end{tabular} 
\caption{Rank list based on weighted average F1-score along with other evaluation metrics (Precision and Recall) for the Malayalam track} 
\label{tab:res_mal} 
\end{center} 
\end{table}

\begin{itemize}
    \item MUCIC \cite{MUCICDravidianCodeMix21}: The authors extracted the character level and syllable level features from the text, which were then used to create the TF-IDF feature vectors. The authors have documented three models, namely: a logistic regression model, an LSTM classifier, and a multilayer perceptron classifier, to classify the messages. The TF-IDF feature vectors are fed to these models, which in turn are trained on the classification task.
    
    \item CIA\_NITT \cite{CIANITTDravidianCodeMix21}: The authors proposed a system that uses a pretrained XLM-RoBERTa for sequence classification. They tokenize the input text using the SentencePiece tokenizer, which is then fed as embeddings to be fine-tuned for the XLM-RoBERTa model.
    .
    \item ZYBank-AI \cite{ZYBankDravidianCodeMix21}: The authors based their experiments on the XLM-RoBERTa as well. To improve the results, they have added self-attention to the 12 hidden layers of the XLMRoBERTA. Furthermore, they propose a two-stage pipeline for the task at hand. In the first stage, the model is trained on data from Dravidian-CodeMix-FIRE 2020. In the second stage, the pre-trained model is fine-tuned on the Dravidian-CodeMix-FIRE 2021 and evaluated on test data.
    
    \item IIITT-Pawan \cite{IIITDravidianCodeMix21}: The authors proposed an ensemble of several fine-tuned language models for sequence classification: BERT, MuRIL, XLM-RoBERTa, DistilBERT. Each of the classifiers is separately trained on training data. During testing, soft voting is employed among all of these classifiers to predict the most likely class.
    \item SOA\_NLP\cite{SOANLPDravidianCodeMix21}: The authors proposed the following two ensemble models for tackling the problem at hand: an ensemble of support vector machine, logistic regression and random forest for Kannada-English texts and an ensemble of support vector machine and logistic regression for    Malayalam-English and Tamil-English texts.
    
    \item SSNCSE\_NLP \cite{SSNCSEDravidianCodeMix21}: The authors have carried out experiments with different features such as  TF-IDF  vectors, count vectorizer and contextual transformer embeddings on primitive machine learning models.
    
  \item IIIT DWD \cite{IIITDWDDravidianCodeMix21}: The authors used pre-trained Word2Vec word embeddings and a parallel RNN model to feed the embeddings into, and have reported their findings on all three datasets.
  
  \item IIIT Surat \cite{IIITSurat_DravidianCodeMix21}: The authors used several feature extraction and preprocessing techniques and then used GLoVe word embeddings and then fed those embeddings to Bi-directional Long-Short Term Memory (Bi-LSTM) model for further processing. For Char embedding, 64 units of Bi-LSTM were used, whereas for processing the words, 32 units of Bi-LSTM was used.
  
  \item SSN\_NLP\_MLRG \cite{SSNMLRGDravidianCodeMix21}: The authors experimented with several machine learning algorithms during the validation process and then fine-tuned the MBERT model to build the system and predict the sentiment polarity for the Tamil-English, Malayalam-English, and Kannada-English languages.
  
  \item IRLab@IITBHU \cite{IRLab_DravidianCodeMix21}: The authors examined if the use of meta embeddings such as FastText will give an edge over pre-trained embeddings such as mBERT. The authors feed meta embeddings into a multiheaded attention based transformer encoder and then over a BiLSTM layer and concatenating it with TF-IDF embeddings to obtain the final outputs.
  
  \item Amrita\_CEN \cite{Amrita_CEN_NLPDravidianCodeMix21}: The authors implemented three architectures: Deep Neural Network (DNN), Bi-LSTM, and finally, Convolution Neural network (CNN) combined to a hybrid model for all the three test sets. Additionally, the authors use a class-weight optimization technique to handle class imbalance.
  
  \item SSNHacML \cite{SSNHacMLDravidianCodeMix21}: The authors proposed an ensemble framework called Ensemble of Convolutional Neural Network and Multi-Head Attention with Bidirectional GRU (ECMAG) to map the code-mixed user comments to their corresponding sentiments. The model has been tested on the Tamil-English Code mixed dataset. The model takes XLMRoberta multilingual sub-word embeddings of the processed text data as input.  
  
 \item MUM \cite{MUMDravidianCodeMix21}: The authors converted the text data into feature vectors and then fed it into a BiLSTM network. The authors submit their predictions to the code-mixed test sets of Kannada, Malayalam, and Tamil.
 
 \item AIML \cite{AIMLDravidianCodeMix21}: The authors extracted character-level features from the text. The dense neural network then uses the extracted features to classify them into different sentiment classes.
 
 \item KBCNMUJAL \cite{KBCNMUJALDravidianCodeMix21}: The authors presented their systems for all three Dravidian Languages (Kannada-English, Tamil-English and Malayalam-English). They use models such as Multinomial Bayes (MNB), CNN and neural networks.
 
 \item Dynamic Duo \cite{DynamicDuoDravidianCodeMix21}: The authors used a  pre-trained language-based Model (BERT), wrapped with ktrain (a python library for model training and testing) to train and validate the data. The authors present their findings on the code-mixed Kannada-English dataset.
 
 \item Ryzer\cite{RYZER_DravidianCodeMix21}: The authors used conventional translation and transliteration algorithms to convert the corpus into a native Tamil script and then fed the data into pre-trained language models like mBert, ULMFit, DistilBert. Additionally, They tested the approach on CNN-BiLSTM and ULMFiT. 
 
 \item SSN\_IT\_NLP \cite{SSNITDravidianCodeMix21}: The authors used a conventional machine learning algorithm. The TF-IDF features are extracted and used for sentiment classification using a Random Forest classifier. 
 
 \item SSNCSE\_NLP \cite{SSNCSE_NLP_aDravidianCodeMix21}: The authors employed a variety of feature extraction techniques and concluded that the count, TF-IDF based vectorization, and multilingual transformer encoding technique performs well on the code-mix polarity labelling task. With these features, and acheived
a weighted F1 score of 0.588 for the Tamil-English task, 0.69 for the Malayalam-English task and 0.63 for the Kannada-English tasks respectively.
\end{itemize}

\section{Evaluation}
The distribution of the sentiment classes is imbalanced in both datasets. This takes into account the varying degrees of importance of each class in the dataset. We used a classification report tool from Scikit learn\footnote{\url{https://scikit-learn.org/stable/modules/generated/sklearn.metrics.classification_report.html}}.
\begin{equation}
\text { Precision }=\frac{T P}{T P+F P} \\
\end{equation}

\begin{equation}
\text { Recall }=\frac{T P}{T P+F N} \\
\end{equation}

\begin{equation}
\text { F-Score }=2 * \frac{\text { Precision } * \text { Recall}}{\text { Precision }+\text { Recall}}
\end{equation}

\begin{equation}
P_{\text {weighted }}=\sum_{i=1}^{L}(P \text { of } i \times \text { Weight of } i)
\end{equation}

\begin{equation}
R_{\text {weighted }}=\sum_{i=1}^{L}(R \text { of } i \times \text { Weight of } i)
\end{equation}

\begin{equation}
F-Score_{\text {weighted }}=\sum_{i=1}^{L}(F-Score \text { of } i \times \text { Weight of } i)
\end{equation}

\section{Results and Discussion}
The sentiment analysis shared task was organized for three languages Tamil, Kannada, and Malayalam.  Overall, there are 120 participants registered for this shared task, yet 22 teams have submitted their working notes for Tamil, 17 for Malayalam, and 15 for Kannada.Table \ref{tab:res_kan},  Table \ref{tab:res_tam}, and Table \ref{tab:res_mal}  show the rank lists of Tamil, Malayalam and Kannada in that order. Most of the submissions submit their systems for the three languages, as specified earlier. Here in this section, we highlight the results of all three languages, which have ranked top positions on the dataset. The results are sorted based on the weighted F1 scores. Most of the teams have used transformer-based models such as BERT, DistilBERT, XLM-RoBERTa or other language models that follow its architecture, in spite of not being pretrained on code-mixed text. Due to the presence of a non-native script in our corpus, the teams got the pre-trained model from the libraries and adopted it for our corpus by fine-tuning. Some teams have used Long Short Term Memory (LSTM) and ULMFiT in their experiments. Also, a few other submissions adopted traditional machine learning algorithms such as Naive Bayes (NB), K-Nearest Neighbors, etc., to solve the problem. 

However, LSTM and traditional machine learning algorithms did not yield good results compared to the
transformer-based models. Out of all the proposed models, XLM-RoBERTa and the transformer-based model produced the best outcomes. Even though many systems with different approaches with F1-score less than the baseline,  we accepted those papers to encourage diverse research methods to solve the problem in Dravidian Languages. Most working notes reported class-wise precision, recall, and F1-score. We used weighted F1 scores as our primary evaluation metric.  

Among the Tamil teams, CIA\_NITT \cite{CIANITTDravidianCodeMix21} got the first position with an F1-score of 0.71. This system achieved 0.71 as the precision and recall score is the same as the F-score. The team from ZYBank-AI \cite{ZYBank-AI} achieved the second position with an F-score of 0.68, lagging the top by 0.03. The top five teams attained an F1 score higher than 0.62. Teams placed in the top positions utilized the transformer-based models for their experiments, particularly XLM-RoBERTa. Contextual embeddings are also found to be effective in this method to reach the top positions. In Malayalam, ZYBank-AI \cite{ZYBank-AI} and CIA\_NITT \cite{CIANITTDravidianCodeMix21} teams switched positions with an F1-score of 080 and 0.75, respectively. Team IIITT-Pawan \cite{IIITDravidianCodeMix21} reached the third position with an F1-score of 0.63 and lagged the top team by only 0.08. According to the Kannada benchmark, CIA\_NITT \cite{CIANITTDravidianCodeMix21} secured the third position while SSNCSE\_NLP \cite{SSNCSEDravidianCodeMix21} and MUCIC \cite{MUCICDravidianCodeMix21} teams reached first and second places in the benchmark, respectively. Also, both teams have used traditional machine learning algorithms such as Logistic Regression, SVM with TF-IDF feature vectors. As we can see, these models have overcome the transformer-based models based on the performance and became the best in the Kannada benchmark.

Table \ref{overall-ranks} shows the overall results and teams that are placed in the top three positions. As we can see,  only one team(CIA\_NITT \cite{CIANITTDravidianCodeMix21}) managed to be in the top 3 systems for the languages, along with achieving the best performance on the code-mixed Tamil dataset. Among the systems submitted during the evaluation period, we observe that the best performing models scored a weighted F1-score of 0.63 in Kannada, 0.80 in Malayalam, and 0.71 in Tamil.

\begin{table}[]
\begin{center} 
\begin{tabular}{|l|l|l|}
\hline
\textbf{Language} & \textbf{Team Name} & \textbf{Rank} \\
\hline
\multirow{3}{*}{Tamil} & CIA\_NITT \cite{CIANITTDravidianCodeMix21} & 1  \\ \cline{2-3} 
 &  ZYBank-AI \cite{ZYBank-AI} & 2  \\ \cline{2-3} 
 &  IIITT-Pawan \cite{IIITDravidianCodeMix21}  &  3  \\  \hline
\multirow{3}{*}{Kannada} & SSNCSE\_NLP \cite{SSNCSEDravidianCodeMix21} & 1  \\ \cline{2-3} 
 &  MUCIC \cite{MUCICDravidianCodeMix21} & 2  \\ \cline{2-3} 
 &  CIA\_NITT \cite{CIANITTDravidianCodeMix21} &  3  \\  \hline
\multirow{3}{*}{Malayalam} & ZYBank-AI Team \cite{ZYBank-AI} & 1  \\ \cline{2-3} 
 &  CIA\_NITT \cite{CIANITTDravidianCodeMix21} & 2  \\ \cline{2-3} 
 &  SOA\_NLP \cite{SOANLPDravidianCodeMix21} &  3  \\  \hline
\end{tabular}
\caption{Overall Results with Top Three Ranks}
\label{overall-ranks}
\end{center}
\end{table}

\section{Conclusion}
We present the results of the sentiment analysis shared task on Tamil, Malayalam, and Kannada.
The dataset used in the shared tasks included code-mixed instances obtained from social media.
Specifically, the dataset was created from Youtube comments following human annotation. Most of the participants fine-tuned pretrained multilingual language models. At the same time, the top-performing systems involved the application of attention layers on the contextualized word embeddings and fine-tuning the models pretrained on the previous edition, DravidianCodeMix-2020's training data.
Results indicate that there is room for improvement in all three languages Tamil, Malayalam, and Kannada.
The increase in the number of participants and the better performance of the systems shows an increase in interest in Dravidian NLP.


\begin{acknowledgments}
This publication is the outcome of the research supported in part by a research grant from Science Foundation Ireland (SFI) under Grant Number SFI/12/RC/2289$\_$P2 (Insight$\_$2), and Irish Research Council grant IRCLA/2017/129 (CARDAMOM-Comparative Deep Models of Language for Minority and Historical Languages).
\end{acknowledgments}

\bibliography{sample-ceur,refs}

\begin{thebibliography}{45}
\expandafter\ifx\csname natexlab\endcsname\relax\def\natexlab#1{#1}\fi
\providecommand{\url}[1]{\texttt{#1}}
\providecommand{\href}[2]{#2}
\providecommand{\path}[1]{#1}
\providecommand{\DOIprefix}{doi:}
\providecommand{\ArXivprefix}{arXiv:}
\providecommand{\URLprefix}{URL: }
\providecommand{\Pubmedprefix}{pmid:}
\providecommand{\doi}[1]{\href{http://dx.doi.org/#1}{\path{#1}}}
\providecommand{\Pubmed}[1]{\href{pmid:#1}{\path{#1}}}
\providecommand{\bibinfo}[2]{#2}
\ifx\xfnm\relax \def\xfnm[#1]{\unskip,\space#1}\fi
\bibitem[{Hande et~al.(2021)Hande, Hegde, Priyadharshini, Ponnusamy, Kumaresan,
  Thavareesan, and Chakravarthi}]{Hande2021BenchmarkingML}
\bibinfo{author}{A.~Hande}, \bibinfo{author}{S.~U. Hegde},
  \bibinfo{author}{R.~Priyadharshini}, \bibinfo{author}{R.~Ponnusamy},
  \bibinfo{author}{P.~K. Kumaresan}, \bibinfo{author}{S.~Thavareesan},
  \bibinfo{author}{B.~R. Chakravarthi},
\newblock \bibinfo{title}{{Benchmarking Multi-Task Learning for Sentiment
  Analysis and Offensive Language Identification in Under-Resourced Dravidian
  Languages}},
\newblock \bibinfo{journal}{ArXiv} \bibinfo{volume}{abs/2108.03867}
  (\bibinfo{year}{2021}).
\bibitem[{Chakravarthi et~al.(2021)Chakravarthi, Priyadharshini, Ponnusamy,
  Kumaresan, Sampath, Thenmozhi, Thangasamy, Nallathambi, and
  McCrae}]{Chakravarthi2021DatasetFI}
\bibinfo{author}{B.~R. Chakravarthi}, \bibinfo{author}{R.~Priyadharshini},
  \bibinfo{author}{R.~Ponnusamy}, \bibinfo{author}{P.~K. Kumaresan},
  \bibinfo{author}{K.~Sampath}, \bibinfo{author}{D.~Thenmozhi},
  \bibinfo{author}{S.~Thangasamy}, \bibinfo{author}{R.~Nallathambi},
  \bibinfo{author}{J.~P. McCrae},
\newblock \bibinfo{title}{{Dataset for Identification of Homophobia and
  Transophobia in Multilingual YouTube Comments}},
\newblock \bibinfo{journal}{ArXiv} \bibinfo{volume}{abs/2109.00227}
  (\bibinfo{year}{2021}).
\bibitem[{Hande et~al.(2020)Hande, Priyadharshini, and
  Chakravarthi}]{hande-etal-2020-kancmd}
\bibinfo{author}{A.~Hande}, \bibinfo{author}{R.~Priyadharshini},
  \bibinfo{author}{B.~R. Chakravarthi},
\newblock \bibinfo{title}{{K}an{CMD}: {K}annada {C}ode{M}ixed dataset for
  sentiment analysis and offensive language detection},
\newblock in: \bibinfo{booktitle}{Proceedings of the Third Workshop on
  Computational Modeling of People's Opinions, Personality, and Emotion's in
  Social Media}, \bibinfo{publisher}{Association for Computational
  Linguistics}, \bibinfo{address}{Barcelona, Spain (Online)},
  \bibinfo{year}{2020}, pp. \bibinfo{pages}{54--63}. \URLprefix
  \url{https://aclanthology.org/2020.peoples-1.6}.
\bibitem[{Hande et~al.(2021)Hande, Puranik, Yasaswini, Priyadharshini,
  Thavareesan, Sampath, Shanmugavadivel, Thenmozhi, and
  Chakravarthi}]{Hande2021OffensiveLI}
\bibinfo{author}{A.~Hande}, \bibinfo{author}{K.~Puranik},
  \bibinfo{author}{K.~Yasaswini}, \bibinfo{author}{R.~Priyadharshini},
  \bibinfo{author}{S.~Thavareesan}, \bibinfo{author}{A.~Sampath},
  \bibinfo{author}{K.~Shanmugavadivel}, \bibinfo{author}{D.~Thenmozhi},
  \bibinfo{author}{B.~R. Chakravarthi},
\newblock \bibinfo{title}{{Offensive Language Identification in Low-resourced
  Code-mixed Dravidian languages using Pseudo-labeling}},
\newblock \bibinfo{journal}{ArXiv} \bibinfo{volume}{abs/2108.12177}
  (\bibinfo{year}{2021}).
\bibitem[{Chakravarthi et~al.(2020{\natexlab{a}})Chakravarthi, Muralidaran,
  Priyadharshini, and McCrae}]{chakravarthi-etal-2020-corpus}
\bibinfo{author}{B.~R. Chakravarthi}, \bibinfo{author}{V.~Muralidaran},
  \bibinfo{author}{R.~Priyadharshini}, \bibinfo{author}{J.~P. McCrae},
\newblock \bibinfo{title}{Corpus creation for sentiment analysis in code-mixed
  {T}amil-{E}nglish text},
\newblock in: \bibinfo{booktitle}{Proceedings of the 1st Joint Workshop on
  Spoken Language Technologies for Under-resourced languages (SLTU) and
  Collaboration and Computing for Under-Resourced Languages (CCURL)},
  \bibinfo{publisher}{European Language Resources association},
  \bibinfo{address}{Marseille, France}, \bibinfo{year}{2020}{\natexlab{a}}, pp.
  \bibinfo{pages}{202--210}. \URLprefix
  \url{https://www.aclweb.org/anthology/2020.sltu-1.28}.
\bibitem[{Chakravarthi et~al.(2020{\natexlab{b}})Chakravarthi, Jose,
  Suryawanshi, Sherly, and McCrae}]{chakravarthi-etal-2020-sentiment}
\bibinfo{author}{B.~R. Chakravarthi}, \bibinfo{author}{N.~Jose},
  \bibinfo{author}{S.~Suryawanshi}, \bibinfo{author}{E.~Sherly},
  \bibinfo{author}{J.~P. McCrae},
\newblock \bibinfo{title}{A sentiment analysis dataset for code-mixed
  {M}alayalam-{E}nglish},
\newblock in: \bibinfo{booktitle}{Proceedings of the 1st Joint Workshop on
  Spoken Language Technologies for Under-resourced languages (SLTU) and
  Collaboration and Computing for Under-Resourced Languages (CCURL)},
  \bibinfo{publisher}{European Language Resources association},
  \bibinfo{address}{Marseille, France}, \bibinfo{year}{2020}{\natexlab{b}}, pp.
  \bibinfo{pages}{177--184}. \URLprefix
  \url{https://www.aclweb.org/anthology/2020.sltu-1.25}.
\bibitem[{Chakravarthi(2020)}]{chakravarthi-2020-hopeedi}
\bibinfo{author}{B.~R. Chakravarthi},
\newblock \bibinfo{title}{{H}ope{EDI}: A multilingual hope speech detection
  dataset for equality, diversity, and inclusion},
\newblock in: \bibinfo{booktitle}{Proceedings of the Third Workshop on
  Computational Modeling of People's Opinions, Personality, and Emotion's in
  Social Media}, \bibinfo{publisher}{Association for Computational
  Linguistics}, \bibinfo{address}{Barcelona, Spain (Online)},
  \bibinfo{year}{2020}, pp. \bibinfo{pages}{41--53}. \URLprefix
  \url{https://aclanthology.org/2020.peoples-1.5}.
\bibitem[{Hande et~al.(2021)Hande, Priyadharshini, Sampath, Thamburaj,
  Chandran, and Chakravarthi}]{Hande2021HopeSD}
\bibinfo{author}{A.~Hande}, \bibinfo{author}{R.~Priyadharshini},
  \bibinfo{author}{A.~Sampath}, \bibinfo{author}{K.~Thamburaj},
  \bibinfo{author}{P.~Chandran}, \bibinfo{author}{B.~R. Chakravarthi},
\newblock \bibinfo{title}{{Hope Speech detection in under-resourced Kannada
  language}},
\newblock \bibinfo{journal}{ArXiv} \bibinfo{volume}{abs/2108.04616}
  (\bibinfo{year}{2021}).
\bibitem[{B and G.~U(2021)}]{SSNCSE_NLP_aDravidianCodeMix21}
\bibinfo{author}{B.~B}, \bibinfo{author}{S.~G.~U},
\newblock \bibinfo{title}{{SSNCSE\_NLP@Dravidian-CodeMix-FIRE2021:Machine
  learning based approach for sentiment analysis on Multilingual Code Mixing
  Text}},
\newblock in: \bibinfo{booktitle}{Working Notes of FIRE 2021 - Forum for
  Information Retrieval Evaluation}, \bibinfo{publisher}{CEUR},
  \bibinfo{year}{2021}.
\bibitem[{Balouchzahi et~al.(2021)Balouchzahi, Shashirekha, and
  Sidorov}]{MUCIC}
\bibinfo{author}{F.~Balouchzahi}, \bibinfo{author}{H.~L. Shashirekha},
  \bibinfo{author}{G.~Sidorov},
\newblock \bibinfo{title}{{MUCIC@Dravidian-CodeMix-FIRE2021:CoSaD- Code-Mixed
  Sentiments Analysis for Dravidian Languages}},
\newblock in: \bibinfo{booktitle}{Working Notes of FIRE 2021 - Forum for
  Information Retrieval Evaluation}, \bibinfo{publisher}{CEUR},
  \bibinfo{year}{2021}.
\bibitem[{Prakash~Babu et~al.(2020)Prakash~Babu, Eswari, and Nimmi}]{CIA_NITT}
\bibinfo{author}{Y.~Prakash~Babu}, \bibinfo{author}{R.~Eswari},
  \bibinfo{author}{K.~Nimmi},
\newblock \bibinfo{title}{{CIA\_NITT@Dravidian-CodeMix-FIRE2020:
  Malayalam-English Code Mixed Sentiment Analysis Using Sentence BERT And
  Sentiment Features}},
\newblock in: \bibinfo{booktitle}{FIRE (Working Notes)}, \bibinfo{year}{2020}.
\bibitem[{Kumar et~al.(2021)Kumar, Saumya, and Singh}]{SOA_NLP}
\bibinfo{author}{A.~Kumar}, \bibinfo{author}{S.~Saumya}, \bibinfo{author}{J.~P.
  Singh},
\newblock \bibinfo{title}{{SOA\_NLP@Dravidian-CodeMix-FIRE2021: An
  ensemble-based model for Sentiment Analysis of Dravidian Code-mixed Social
  Media Posts}},
\newblock in: \bibinfo{booktitle}{Working Notes of FIRE 2021 - Forum for
  Information Retrieval Evaluation}, \bibinfo{publisher}{CEUR},
  \bibinfo{year}{2021}.
\bibitem[{Puranik et~al.(2021)Puranik, B, and Kumar}]{IIITT}
\bibinfo{author}{K.~Puranik}, \bibinfo{author}{B.~B}, \bibinfo{author}{B.~S.
  Kumar},
\newblock \bibinfo{title}{{IIITT@Dravidian-CodeMix-FIRE2021: Transliterate or
  translate? Sentiment analysis of code-mixed text in Dravidian languages}},
\newblock in: \bibinfo{booktitle}{Working Notes of FIRE 2021 - Forum for
  Information Retrieval Evaluation}, \bibinfo{publisher}{CEUR},
  \bibinfo{year}{2021}.
\bibitem[{Dutta et~al.(2021)Dutta, Agrawal, and Roy}]{DynamicDuo}
\bibinfo{author}{S.~Dutta}, \bibinfo{author}{H.~Agrawal},
  \bibinfo{author}{P.~K. Roy},
\newblock \bibinfo{title}{{DynamicDuo@Dravidian-CodeMix-FIRE2021: Sentiment
  Analysis on Multilingual Code Mixing Text using BERT}},
\newblock in: \bibinfo{booktitle}{Working Notes of FIRE 2021 - Forum for
  Information Retrieval Evaluation}, \bibinfo{publisher}{CEUR},
  \bibinfo{year}{2021}.
\bibitem[{Pathak et~al.(2020)Pathak, Joshi, Joshi, Mundada, and
  Joshi}]{KBCNMUJAL}
\bibinfo{author}{V.~Pathak}, \bibinfo{author}{M.~Joshi},
  \bibinfo{author}{P.~Joshi}, \bibinfo{author}{M.~Mundada},
  \bibinfo{author}{T.~Joshi},
\newblock \bibinfo{title}{{KBCNMUJAL@HASOC-Dravidian-CodeMix-FIRE2020: Using
  Machine Learning for Detection of Hate Speech and Offensive Codemix Social
  Media text}},
\newblock in: \bibinfo{booktitle}{FIRE (Working Notes)}, \bibinfo{year}{2020}.
\bibitem[{Jada et~al.(2021)Jada, Reddy, Yasaswini, K, Chandran, Sampath, and
  Thangasamy}]{IIIT}
\bibinfo{author}{P.~K. Jada}, \bibinfo{author}{D.~S. Reddy},
  \bibinfo{author}{K.~Yasaswini}, \bibinfo{author}{A.~P. K},
  \bibinfo{author}{P.~Chandran}, \bibinfo{author}{A.~Sampath},
  \bibinfo{author}{S.~Thangasamy},
\newblock \bibinfo{title}{{IIIT@Dravidian-CodeMix-FIRE2021: Transformer based
  Sentiment Analysis in Dravidian Languages}},
\newblock in: \bibinfo{booktitle}{Working Notes of FIRE 2021 - Forum for
  Information Retrieval Evaluation}, \bibinfo{publisher}{CEUR},
  \bibinfo{year}{2021}.
\bibitem[{Kalaivani and Thenmozhi(2020)}]{SSN_NLP_MLRG}
\bibinfo{author}{A.~Kalaivani}, \bibinfo{author}{D.~Thenmozhi},
\newblock \bibinfo{title}{{SSN\_NLP\_MLRG@Dravidian-CodeMix-FIRE2020: Sentiment
  Code-Mixed Text Classification in Tamil and Malayalam using ULMFiT}},
\newblock in: \bibinfo{booktitle}{FIRE (Working Notes)}, \bibinfo{year}{2020}.
\bibitem[{P.H.V et~al.(2021)P.H.V, B, Jp, and Kp}]{Amrita_CEN_NLP}
\bibinfo{author}{P.~K. P.H.V}, \bibinfo{author}{P.~B}, \bibinfo{author}{S.~Jp},
  \bibinfo{author}{S.~Kp},
\newblock \bibinfo{title}{{Amrita\_CEN\_NLP@Dravidian-CodeMix-FIRE2021:Deep
  Learning based Sentiment analysis forMalayalam,Tamil and Kannada languages
  }},
\newblock in: \bibinfo{booktitle}{Working Notes of FIRE 2021 - Forum for
  Information Retrieval Evaluation}, \bibinfo{publisher}{CEUR},
  \bibinfo{year}{2021}.
\bibitem[{M~D and H~L(2021)}]{MUM}
\bibinfo{author}{A.~M~D}, \bibinfo{author}{S.~H~L},
\newblock \bibinfo{title}{{MUM@Dravidian-CodeMix-FIRE2021:BiLSTM-Sentiments
  Analysis in Code MixedDravidian Languages}},
\newblock in: \bibinfo{booktitle}{Working Notes of FIRE 2021 - Forum for
  Information Retrieval Evaluation}, \bibinfo{publisher}{CEUR},
  \bibinfo{year}{2021}.
\bibitem[{Saroj and Pal(2020)}]{IRLab}
\bibinfo{author}{A.~Saroj}, \bibinfo{author}{S.~Pal},
\newblock \bibinfo{title}{{IRLab@IIT-BHU@Dravidian-CodeMix-FIRE2020: Sentiment
  Analysis on Multilingual Code Mixing Text Using BERT-{BASE}}},
\newblock in: \bibinfo{booktitle}{FIRE (Working Notes)}, \bibinfo{year}{2020}.
\bibitem[{Bai et~al.(2021)Bai, Zhang, Gu, Guan, and Shi}]{ZYBank-AI}
\bibinfo{author}{Y.~Bai}, \bibinfo{author}{B.~Zhang}, \bibinfo{author}{Y.~Gu},
  \bibinfo{author}{T.~Guan}, \bibinfo{author}{Q.~Shi},
\newblock \bibinfo{title}{{ZYBank-AI@Dravidian-CodeMix-FIRE2021: Automatic
  Detecting the Sentiment of Code-Mixed Text by Pre-training Model }},
\newblock in: \bibinfo{booktitle}{Working Notes of FIRE 2021 - Forum for
  Information Retrieval Evaluation}, \bibinfo{publisher}{CEUR},
  \bibinfo{year}{2021}.
\bibitem[{Sivapiran et~al.(2021)Sivapiran, Vasantharajan, and
  Thayasivam}]{RYZER}
\bibinfo{author}{S.~Sivapiran}, \bibinfo{author}{C.~Vasantharajan},
  \bibinfo{author}{U.~Thayasivam},
\newblock \bibinfo{title}{{RYZER@Dravidian-CodeMix-FIRE2021: Sentiment Analysis
  in Dravidian Code-Mixed YouTube Comments and Posts }},
\newblock in: \bibinfo{booktitle}{Working Notes of FIRE 2021 - Forum for
  Information Retrieval Evaluation}, \bibinfo{publisher}{CEUR},
  \bibinfo{year}{2021}.
\bibitem[{Kumari and Kumar(2021)}]{AIML}
\bibinfo{author}{J.~Kumari}, \bibinfo{author}{A.~Kumar},
\newblock \bibinfo{title}{{AIML@Dravidian-CodeMix-FIRE2021: A Deep Neural
  Network-based Model for the Sentiment Analysis of Dravidian Code-mixed Social
  Media Posts }},
\newblock in: \bibinfo{booktitle}{Working Notes of FIRE 2021 - Forum for
  Information Retrieval Evaluation}, \bibinfo{publisher}{CEUR},
  \bibinfo{year}{2021}.
\bibitem[{Mishra et~al.(2021)Mishra, Saumya, and Kumar}]{IIIT_DWD}
\bibinfo{author}{A.~K. Mishra}, \bibinfo{author}{S.~Saumya},
  \bibinfo{author}{A.~Kumar},
\newblock \bibinfo{title}{{IIIT\_DWD@HASOC 2021:Sentiment analysis of
  Dravidian-CodeMix language }},
\newblock in: \bibinfo{booktitle}{Working Notes of FIRE 2021 - Forum for
  Information Retrieval Evaluation}, \bibinfo{publisher}{CEUR},
  \bibinfo{year}{2021}.
\bibitem[{Roy and Kumar(2021)}]{IIITSurat}
\bibinfo{author}{P.~K. Roy}, \bibinfo{author}{A.~Kumar},
\newblock \bibinfo{title}{{IIITSurat@Dravidian-CodeMix-FIRE2021: Sentiment
  Analysis on Tamil Code-Mixed Text using Bi-LSTM }},
\newblock in: \bibinfo{booktitle}{Working Notes of FIRE 2021 - Forum for
  Information Retrieval Evaluation}, \bibinfo{publisher}{CEUR},
  \bibinfo{year}{2021}.
\bibitem[{D et~al.(2021)D, J~B, and Durairaj}]{SSNHacML}
\bibinfo{author}{P.~D}, \bibinfo{author}{S.~J~B},
  \bibinfo{author}{T.~Durairaj},
\newblock \bibinfo{title}{{ SSNHacML@Dravidian-CodeMix-FIRE2021: ECMAG -
  Ensemble of CNN and Multi-Head Attention with Bi-GRU for Sentiment Analysis
  in Code-Mixed Data}},
\newblock in: \bibinfo{booktitle}{Working Notes of FIRE 2021 - Forum for
  Information Retrieval Evaluation}, \bibinfo{publisher}{CEUR},
  \bibinfo{year}{2021}.
\bibitem[{N and S(2021)}]{SSN_IT_NLP}
\bibinfo{author}{S.~N}, \bibinfo{author}{D.~S},
\newblock \bibinfo{title}{{SSN\_IT\_NLP@Dravidian-CodeMix-FIRE2021: Opinion And
  Attitude Investigation }},
\newblock in: \bibinfo{booktitle}{Working Notes of FIRE 2021 - Forum for
  Information Retrieval Evaluation}, \bibinfo{publisher}{CEUR},
  \bibinfo{year}{2021}.
\bibitem[{Balouchzahi et~al.(2021)Balouchzahi, Shashirekha, and
  Sidorov}]{MUCICDravidianCodeMix21}
\bibinfo{author}{F.~Balouchzahi}, \bibinfo{author}{H.~L. Shashirekha},
  \bibinfo{author}{G.~Sidorov},
\newblock \bibinfo{title}{{MUCIC@Dravidian-CodeMix-FIRE2021:CoSaD- Code-Mixed
  Sentiments Analysis for Dravidian Languages}},
\newblock in: \bibinfo{booktitle}{Working Notes of FIRE 2021 - Forum for
  Information Retrieval Evaluation}, \bibinfo{publisher}{CEUR},
  \bibinfo{year}{2021}.
\bibitem[{Prakash~Babu and Eswari(2021)}]{CIANITTDravidianCodeMix21}
\bibinfo{author}{Y.~Prakash~Babu}, \bibinfo{author}{R.~Eswari},
\newblock \bibinfo{title}{{CIA\_NITT@Dravidian-CodeMix-FIRE2021:Sentiment
  Analysis on Dravidian Code-Mixed YouTube Comments using Paraphrase XLMRoBERTa
  Model}},
\newblock in: \bibinfo{booktitle}{Working Notes of FIRE 2021 - Forum for
  Information Retrieval Evaluation}, \bibinfo{publisher}{CEUR},
  \bibinfo{year}{2021}.
\bibitem[{Bai et~al.(2020)Bai, Zhang, Gu, Guan, and
  Shi}]{ZYBankDravidianCodeMix21}
\bibinfo{author}{Y.~B. Bai}, \bibinfo{author}{B.~Zhang},
  \bibinfo{author}{Y.~Gu}, \bibinfo{author}{T.~Guan}, \bibinfo{author}{Q.~Shi},
\newblock \bibinfo{title}{{ZYBank-AI@Dravidian-CodeMix-FIRE2021: Automatic
  Detecting the Sentiment of Code-Mixed Text by Pre-training Model}},
\newblock in: \bibinfo{booktitle}{FIRE (Working Notes)}, \bibinfo{year}{2020}.
\bibitem[{Jadaa et~al.(2020)Jadaa, Reddy, Yasawini, Pandian~K, Chandran,
  Sampath, and Thangasamy}]{IIITDravidianCodeMix21}
\bibinfo{author}{P.~K. Jadaa}, \bibinfo{author}{S.~Reddy},
  \bibinfo{author}{K.~Yasawini}, \bibinfo{author}{A.~Pandian~K},
  \bibinfo{author}{P.~Chandran}, \bibinfo{author}{A.~Sampath},
  \bibinfo{author}{S.~Thangasamy},
\newblock \bibinfo{title}{{IIITT@Dravidian-CodeMix-FIRE2021: Transformer based
  Sentiment Analysis in Dravidian Languages}},
\newblock in: \bibinfo{booktitle}{FIRE (Working Notes)}, \bibinfo{year}{2020}.
\bibitem[{Kumar et~al.(2020)Kumar, Saumya, and
  Singh}]{SOANLPDravidianCodeMix21}
\bibinfo{author}{A.~Kumar}, \bibinfo{author}{S.~Saumya}, \bibinfo{author}{J.~P.
  Singh},
\newblock \bibinfo{title}{{SOA\_NLP@Dravidian-CodeMix-FIRE2021: An
  ensemble-based model for Sentiment Analysis of Dravidian Code-mixed Social
  Media Posts}},
\newblock in: \bibinfo{booktitle}{FIRE (Working Notes)}, \bibinfo{year}{2020}.
\bibitem[{Bharathi and Samyuktha(2020)}]{SSNCSEDravidianCodeMix21}
\bibinfo{author}{B.~Bharathi}, \bibinfo{author}{G.~Samyuktha},
\newblock \bibinfo{title}{{SSNCSE\_NLP@Dravidian-CodeMix-FIRE2021: Machine
  learning based approach for sentiment Analysis on Multilingual Code Mixing
  Text}},
\newblock in: \bibinfo{booktitle}{FIRE (Working Notes)}, \bibinfo{year}{2020}.
\bibitem[{Mishra et~al.(2021)Mishra, Saumya, and
  Kumar}]{IIITDWDDravidianCodeMix21}
\bibinfo{author}{A.~K. Mishra}, \bibinfo{author}{S.~Saumya},
  \bibinfo{author}{A.~Kumar},
\newblock \bibinfo{title}{{IIIT\_DWD@Dravidian-CodeMix-FIRE2021:Sentiment
  analysis of Dravidian-CodeMix language}},
\newblock in: \bibinfo{booktitle}{Working Notes of FIRE 2021 - Forum for
  Information Retrieval Evaluation}, \bibinfo{publisher}{CEUR},
  \bibinfo{year}{2021}.
\bibitem[{Roy and Kumar(2021)}]{IIITSurat_DravidianCodeMix21}
\bibinfo{author}{P.~K. Roy}, \bibinfo{author}{A.~Kumar},
\newblock \bibinfo{title}{{IIITSurat@Dravidian-CodeMix-FIRE2021: Sentiment
  Analysis on Tamil Code-Mixed Text using Bi-LSTM}},
\newblock in: \bibinfo{booktitle}{Working Notes of FIRE 2021 - Forum for
  Information Retrieval Evaluation}, \bibinfo{publisher}{CEUR},
  \bibinfo{year}{2021}.
\bibitem[{Adaikkan and Durairaj(2021)}]{SSNMLRGDravidianCodeMix21}
\bibinfo{author}{K.~Adaikkan}, \bibinfo{author}{T.~Durairaj},
\newblock \bibinfo{title}{{SSN\_NLP\_MLRG@Dravidian-CodeMix-FIRE2021:
  Multilingual Sentiment Analysis in Tamil, Malayalam, and Kannada code-mixed
  social media posts}},
\newblock in: \bibinfo{booktitle}{Working Notes of FIRE 2021 - Forum for
  Information Retrieval Evaluation}, \bibinfo{publisher}{CEUR},
  \bibinfo{year}{2021}.
\bibitem[{Chanda et~al.(2021)Chanda, Singh, and Pal}]{IRLab_DravidianCodeMix21}
\bibinfo{author}{S.~Chanda}, \bibinfo{author}{R.~P. Singh},
  \bibinfo{author}{S.~Pal},
\newblock \bibinfo{title}{{IRLab@IITBHU@Dravidian-CodeMix-FIRE2021: Is Meta
  Embedding better than pre-trained word embedding to perform Sentiment
  Analysis for Dravidian Languages in Code-Mixed Text?}},
\newblock in: \bibinfo{booktitle}{Working Notes of FIRE 2021 - Forum for
  Information Retrieval Evaluation}, \bibinfo{publisher}{CEUR},
  \bibinfo{year}{2021}.
\bibitem[{Kumar et~al.(2021)Kumar, B, J.P, and
  KP}]{Amrita_CEN_NLPDravidianCodeMix21}
\bibinfo{author}{P.~Kumar}, \bibinfo{author}{P.~B}, \bibinfo{author}{S.~J.P},
  \bibinfo{author}{S.~KP},
\newblock \bibinfo{title}{{Amrita\_CEN\_NLP@Dravidian-CodeMix-FIRE2021:Deep
  Learning based Sentiment analysis for Malayalam,Tamil and Kannada
  languages}},
\newblock in: \bibinfo{booktitle}{Working Notes of FIRE 2021 - Forum for
  Information Retrieval Evaluation}, \bibinfo{publisher}{CEUR},
  \bibinfo{year}{2021}.
\bibitem[{D et~al.(2021)D, J~B, and Durairaj}]{SSNHacMLDravidianCodeMix21}
\bibinfo{author}{P.~D}, \bibinfo{author}{S.~J~B},
  \bibinfo{author}{T.~Durairaj},
\newblock \bibinfo{title}{{SSNHacML@Dravidian-CodeMix-FIRE2021: ECMAG -
  Ensemble of CNN and Multi-Head Attention with Bi-GRU for Sentiment Analysis
  in Code-Mixed Data}},
\newblock in: \bibinfo{booktitle}{Working Notes of FIRE 2021 - Forum for
  Information Retrieval Evaluation}, \bibinfo{publisher}{CEUR},
  \bibinfo{year}{2021}.
\bibitem[{MD and Shashirekha(2021)}]{MUMDravidianCodeMix21}
\bibinfo{author}{A.~MD}, \bibinfo{author}{H.~L. Shashirekha},
\newblock \bibinfo{title}{{MUM@Dravidian-CodeMix-FIRE2021:BiLSTM-Sentiments
  Analysis in Code MixedDravidian Languages}},
\newblock in: \bibinfo{booktitle}{Working Notes of FIRE 2021 - Forum for
  Information Retrieval Evaluation}, \bibinfo{publisher}{CEUR},
  \bibinfo{year}{2021}.
\bibitem[{Kumari and Kumar(2021)}]{AIMLDravidianCodeMix21}
\bibinfo{author}{J.~Kumari}, \bibinfo{author}{A.~Kumar},
\newblock \bibinfo{title}{{AIML@Dravidian-CodeMix-FIRE2021: A Deep Neural
  Network-based Model for the Sentiment Analysis of Dravidian Code-mixed Social
  Media Posts}},
\newblock in: \bibinfo{booktitle}{Working Notes of FIRE 2021 - Forum for
  Information Retrieval Evaluation}, \bibinfo{publisher}{CEUR},
  \bibinfo{year}{2021}.
\bibitem[{Joshi and Pathak(2021)}]{KBCNMUJALDravidianCodeMix21}
\bibinfo{author}{P.~Joshi}, \bibinfo{author}{V.~Pathak},
\newblock \bibinfo{title}{{KBCNMUJAL@Dravidian-CodeMix-HASOC2021:Offensive
  Language Identification on Code-mixed Dravidian Languages, A Non-linguistic
  Approach}},
\newblock in: \bibinfo{booktitle}{Working Notes of FIRE 2021 - Forum for
  Information Retrieval Evaluation}, \bibinfo{publisher}{CEUR},
  \bibinfo{year}{2021}.
\bibitem[{Dutta et~al.(2021)Dutta, Agrawal, and
  Roy}]{DynamicDuoDravidianCodeMix21}
\bibinfo{author}{S.~Dutta}, \bibinfo{author}{H.~Agrawal},
  \bibinfo{author}{P.~K. Roy},
\newblock \bibinfo{title}{{DynamicDuo@Dravidian-CodeMix-FIRE2021: Sentiment
  Analysis on Multilingual Code Mixing Text using BERT}},
\newblock in: \bibinfo{booktitle}{Working Notes of FIRE 2021 - Forum for
  Information Retrieval Evaluation}, \bibinfo{publisher}{CEUR},
  \bibinfo{year}{2021}.
\bibitem[{Sivapiran et~al.(2021)Sivapiran, Vasantharajan, and
  Thayasivam}]{RYZER_DravidianCodeMix21}
\bibinfo{author}{S.~Sivapiran}, \bibinfo{author}{C.~Vasantharajan},
  \bibinfo{author}{U.~Thayasivam},
\newblock \bibinfo{title}{{RYZER@Dravidian-CodeMix-FIRE2021: Sentiment Analysis
  in Dravidian Code-Mixed YouTube Comments and Posts}},
\newblock in: \bibinfo{booktitle}{Working Notes of FIRE 2021 - Forum for
  Information Retrieval Evaluation}, \bibinfo{publisher}{CEUR},
  \bibinfo{year}{2021}.
\bibitem[{N and S(2021)}]{SSNITDravidianCodeMix21}
\bibinfo{author}{S.~N}, \bibinfo{author}{D.~S},
\newblock \bibinfo{title}{{SSN\_IT\_NLP@Dravidian-CodeMix-FIRE2021: Opinion And
  Attitude Investigation}},
\newblock in: \bibinfo{booktitle}{Working Notes of FIRE 2021 - Forum for
  Information Retrieval Evaluation}, \bibinfo{publisher}{CEUR},
  \bibinfo{year}{2021}.

\end{thebibliography}

\end{document}